\pdfoutput=1
\documentclass[runningheads]{llncs}
\usepackage{graphicx}
\usepackage{enumitem}
\usepackage{tabu}
\usepackage{booktabs}
\usepackage[outercaption]{sidecap}  
\usepackage{multirow}

\usepackage{amsmath,amssymb} 
\usepackage{color}
\makeatletter
\newcommand{\printfnsymbol}[1]{%
	\textsuperscript{\@fnsymbol{#1}}%
}
\makeatother
\begin{document}
\title{Synthetically Trained Icon Proposals\\ for Parsing and Summarizing Infographics} 

\titlerunning{Synthetically Trained Icon Proposals}
%
\author{Spandan Madan\thanks{Authors contributed equally to this work.}\inst{1} \and
Zoya Bylinskii\printfnsymbol{1} \inst{1} \and
Matthew Tancik\printfnsymbol{1} \inst{1}\and\\
Adri\`{a} Recasens\inst{1}\and
Kimberli Zhong \inst{1}\and
Sami Alsheikh \inst{1}\and
Hanspeter Pfister \inst{2}\and
Aude Oliva \inst{1}\and
Fredo Durand\inst{1}}
%
\authorrunning{Madan et al.}
%

\institute{Massachusetts Institute of Technology, Cambridge, MA, USA\\
\email{\{smadan,zoya,tancik,recasens,kimberli,alsheikh,oliva,fredo\}@mit.edu}\\ \and
Harvard University, Cambridge, MA, USA\\
\email{pfister@seas.harvard.edu}}
\maketitle              
\begin{abstract}
Widely used in news, business, and educational media, infographics are handcrafted to effectively communicate messages about complex and often abstract topics including `ways to conserve the environment' and `understanding the financial crisis'. Composed of stylistically and semantically diverse visual and textual elements, infographics pose new challenges for computer vision. While automatic text extraction works well on infographics, computer vision approaches trained on natural images fail to identify the stand-alone visual elements in infographics, or `icons'. To bridge this representation gap, we propose a synthetic data generation strategy: we augment background patches in infographics from our Visually29K dataset with Internet-scraped icons which we use as training data for an icon proposal mechanism. On a test set of 1K annotated infographics, icons are located with 38\% precision and 34\% recall (the best model trained with natural images achieves 14\% precision and 7\% recall). Combining our icon proposals with icon classification and text extraction, we present a multi-modal summarization application. Our application takes an infographic as input and automatically produces text tags and visual hashtags that are textually and visually representative of the infographic's topics respectively.

\end{abstract}
\section{Introduction}
A vast amount of semantic and design knowledge is encoded in graphic designs, 
which are created to effectively communicate messages about complex and often abstract topics.
Graphic designs include clipart~\cite{zitnick2013bringing,zitnick2016adopting}, comics~\cite{comicspaper}, advertisements~\cite{hussain2017automatic,wilber2017bam}, diagrams~\cite{kembhavi2016eccv,seo2014diagram}, and the infographics that are the focus of this paper. 
Expanding the capabilities of computational algorithms to understand graphic designs can therefore complement natural image understanding 
by motivating a set of novel research questions with unique challenges. In particular, current techniques trained for natural images do not generalize to the abstract visual elements and diverse styles in graphic designs~\cite{comicspaper,hussain2017automatic,wilber2017bam}. 
\begin{figure}
	\centering
	\includegraphics[width=1\linewidth]{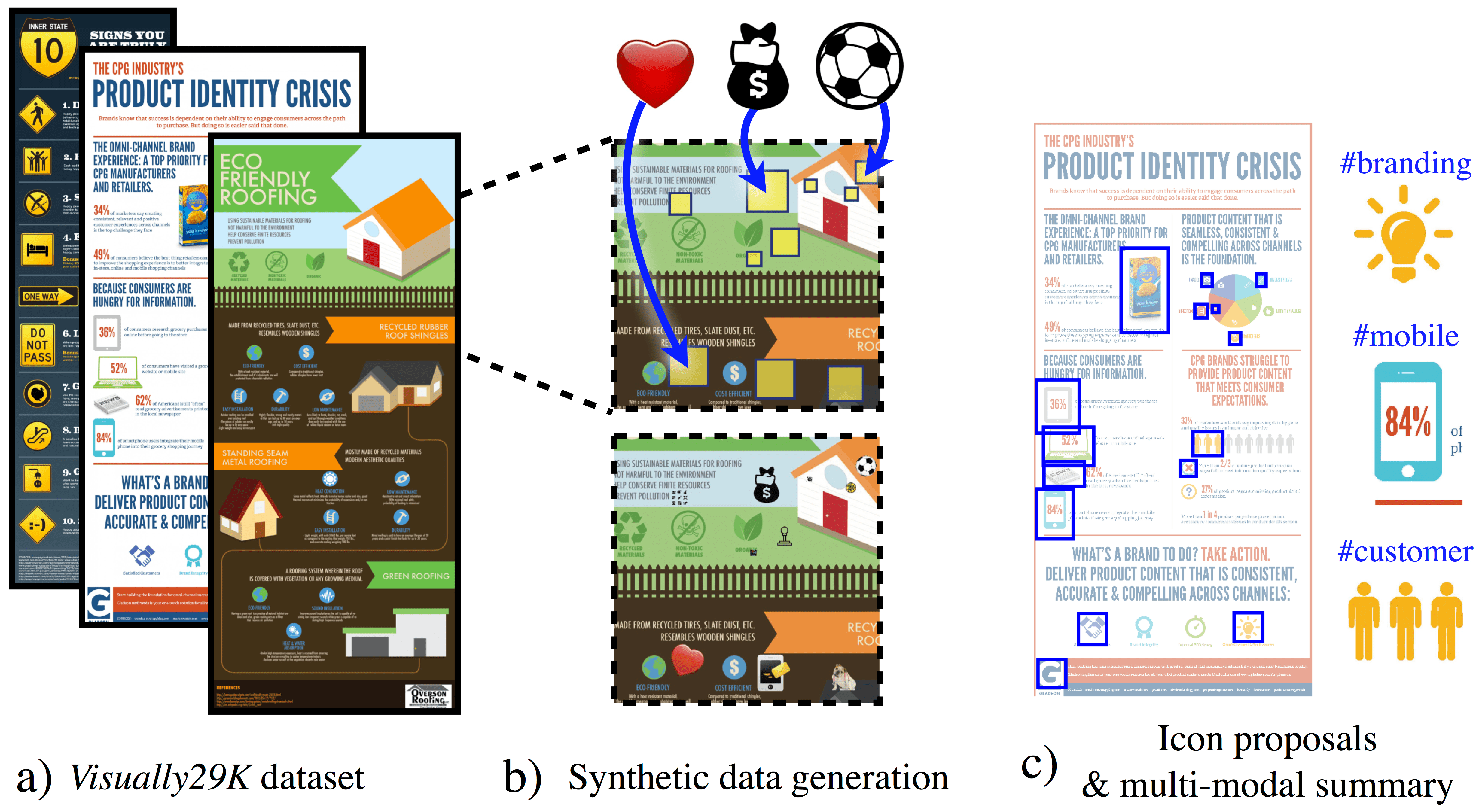}
	\caption{We make 3 contributions: a) We present \emph{Visually29K}, a curated dataset of infographics; b) We generate synthetic data by augmenting Internet-scraped icons onto patches of infographics to train an icon proposal mechanism; c) We evaluate our automatic icon proposals and present a multi-modal summarization application that takes an infographic and outputs the text tags and visual hashtags that are most representative of the infographic's topics.}
	\label{fig:multimodalexamples}
\end{figure}

\begin{SCfigure}
	\includegraphics[width=0.6\linewidth]{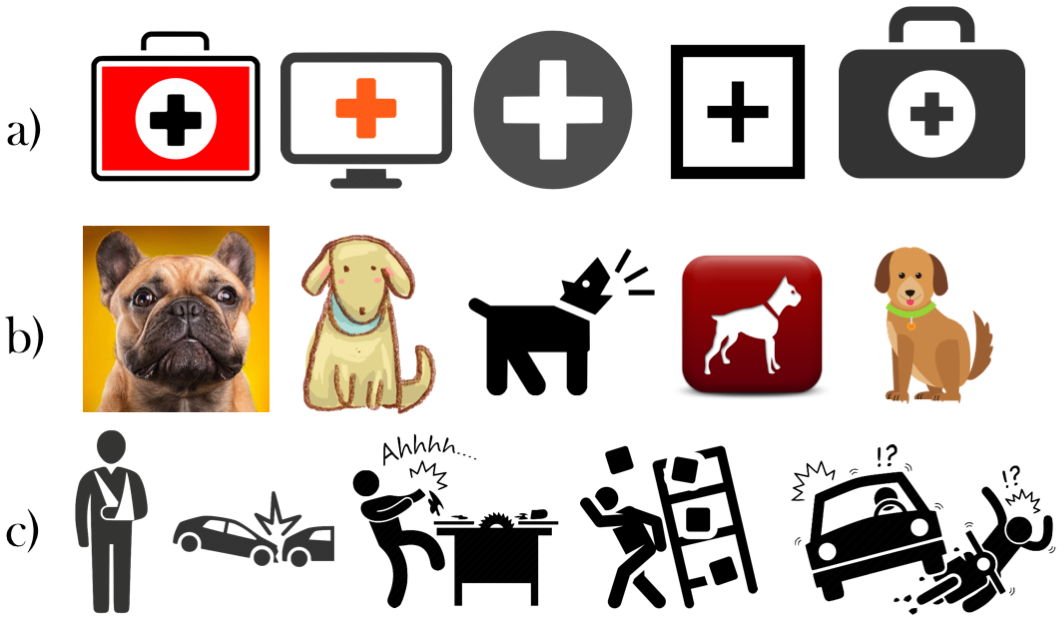}
	\caption{Examples of stylistic and semantic variations in scraped icons. a) Visually similar icons scraped for different but semantically related tags \emph{medical, doctor, health, hospital, medicine}. b)~Icons with varied styles scraped for the tag \emph{dog}. c) Icons with varied semantic representations for the tag \emph{accident}.}
	\label{fig:scraped_icons}
\end{SCfigure}

In this paper, we tackle the challenge of identifying stand-alone visual elements, which we call `icons'. Instead of (class-specific) icon detection, we instead wish to locate all icon-like elements in an image - i.e., to generate icon proposals.
To adapt to the stylistic, semantic, and scale variations of icons in graphic designs (Fig.~\ref{fig:scraped_icons}), which differentiate them from objects in natural images, we propose a synthetic data generation approach. We augment background patches in infographics with a dataset of Internet-scraped icons which we use  
as training data for an icon proposal mechanism (Fig.~\ref{fig:multimodalexamples}b). Our resulting icon proposals outperform models trained on natural images, achieving 38\% precision and 34\% recall (YOLO9000~\cite{YOLO9000}, trained on ImageNet~\cite{Imagenet}, reaches 14\% precision and 7\% recall), showing that a representation gap exists between objects in natural images and icons in graphic designs.

For training computational models, we curated a novel dataset of 29K infographics from the \emph{Visual.ly} design website (Fig.~\ref{fig:multimodalexamples}a). Infographics are a form of graphic designs widely used in news, business, and educational media to cover diverse topics, including `ways to conserve the environment' and `understanding the financial crisis'.
Each infographic in our dataset is annotated with 1-9 tags, out of a set of 391 tag categories. For 1,400 infographics, we collected a total of 21,288 human-annotated bounding boxes of icon locations. For another subset of 544 infographics, we collected 7,761 tagged (categorized) icon bounding boxes. We analyzed human consistency and used the annotations to evaluate our automatic approaches. 

Finally, to showcase an example application that makes use of icons, we use our automatic icon proposals in combination with icon classification and text extraction to present a novel multi-modal summarization application on our infographics dataset (Fig.~\ref{fig:multimodalexamples}c). Given an infographic as input, our application automatically outputs text tags and visual hashtags that are textually and visually representative of the infographic's topics, respectively. This presents a first step towards combining the textual and visual information in an infographic for computational understanding. Together with existing methods for text extraction, our automatic icon proposals can facilitate future applications including knowledge retrieval, visual search, captioning, and visual question answering.

\textbf{Contributions:} In this paper, we introduce: (a) \emph{Visually29K}, a novel dataset of infographics that we will make publicly available; (b) a synthetic data generation strategy to train an icon proposal mechanism; (c) a proposed multi-modal summarization application, using detected icons and text to automatically output representative text tags and \emph{visual hashtags} for infographics.
\section{Related Work}


\textbf{Computer vision for graphic designs:} 
Computer vision has traditionally focused on understanding natural images and scenes. However, there is a growing interest in graphic designs, which motivates a new set of research questions and technical challenges. Zitnick et al.~\cite{zitnick2013bringing,zitnick2016adopting} introduced abstract scenes to study higher-level image semantics (relationships between objects, storylines, etc.).
Wilber et al.~\cite{wilber2017bam} presented an `Artistic Media Dataset' to explore the representation gap between objects in photographs versus in artistic media. Iyyer et al.~\cite{comicspaper} built a `COMICS' dataset and made predictions about actions and characters using extracted visual and textual elements from comic panels. Hussain et al.~\cite{hussain2017automatic} presented a dataset of advertisements and described the challenges of parsing symbolism, memes, humor, and physical properties from images.
Computer vision tools have also been used to transcribe textbook diagrams into structured tables for question answering \cite{kembhavi2016eccv,seo2014diagram}, and to parse graphs and charts for retargeting applications \cite{bylinskii2017learning,2017-reverse-engineering-vis,2011-revision}. 
To the best of our knowledge, there is no work on automated understanding of infographics or using computer vision techniques to identify icons in graphic designs. 

\textbf{Synthetic training data:}
The use of synthetically generated data to train large CNN models has been gaining popularity, e.g., for learning optical flow~\cite{Sintel}, action recognition~\cite{PHAV}, overcoming scattering~\cite{satat2017object}, and object tracking~\cite{VirtualKITTI}. Simulated environments like video games have been used to collect realistic scene images for semantic segmentation~\cite{VideoGame}. Our work was inspired by a text recognition system which was trained on a synthetic dataset of images augmented with text~\cite{oxford_textspotter}. 
Related to our approach, Dwibedi et al.~\cite{dwibedi2017cut} insert segmented objects into real images to learn to detect natural objects in the wild. We leverage the fact that infographics are digitally-born, so augmenting them with more Internet-scraped design elements is a natural step. Tsutsui and Crandall~\cite{tsutsui2017data} synthesize compound figures by randomly arranging them on white backgrounds to learn to re-detect them. However, the icons we aim to detect occur on top of complex backgrounds, so we need our synthetic data to capture the visual statistics of in-the-wild infographics (Sec.~\ref{sec:icondataset}).  
\section{Visually29K: an infographics dataset}\label{sec:visually}
To facilitate computer vision research on infographics,
we assembled the \emph{Visually29K} dataset. We scraped 63K static infographic images from the \emph{Visual.ly} website, a community platform for human-designed visual content.
Each infographic is hand categorized, tagged, and described by a designer, making it a rich source of annotated data. 
We curated this dataset to obtain a representative subset of 28,973 images, ensuring sufficient instances per tag (Table~\ref{tab:datasetstats}). The tags associated with images are free-form text, so many of the original tags were either semantically redundant or had too few instances. We cut the original heavy-tailed distribution of 19K tags down to 391 tags with at least 50 exemplars, and by merging redundant tags manually using WordNet~\cite{miller1995wordnet}. 
Tags range from concepts which have concrete visual depictions (e.g., \emph{car, cat, baby}) to abstract concepts (e.g., \emph{search engine optimization, foreclosure, revenue}). 
Metadata for this dataset also includes labels for 26 categories (available for 90\% of the infographics), titles (99\%) and descriptions (94\%), available for future applications.

\begin{SCtable}
	\small
	\begin{tabular}{c c c c c}
		\hline
		\textbf{Dataset} & \begin{tabular}[c]{@{}c@{}}\textbf{\# of}\\ \textbf{tags}\end{tabular} & \begin{tabular}[c]{@{}c@{}}\textbf{Images}\\ \textbf{per tag}\end{tabular}              & \begin{tabular}[c]{@{}c@{}}\textbf{Tags per}\\ \textbf{Image}\end{tabular}
		& \begin{tabular}[c]{@{}c@{}}\textbf{Aspect}\\ \textbf{ratios}\end{tabular}
		\\ \hline
		\begin{tabular}[c]{@{}c@{}}63K\\ (full)\end{tabular}  & 19469                                               & \begin{tabular}[c]{@{}c@{}}min=1\\ max=3784\\ mean=7.8\end{tabular}  & \begin{tabular}[c]{@{}c@{}}min=0\\ max=10\\ mean=3.7\end{tabular}  & \begin{tabular}[c]{@{}c@{}} min=1:20\\ max=22:1 \end{tabular}\\ \hline
		\begin{tabular}[c]{@{}c@{}}29K\\ (curated)\end{tabular} & 391                                                  & \begin{tabular}[c]{@{}c@{}}min=50\\ max=2331\\ mean=151\end{tabular} & \begin{tabular}[c]{@{}c@{}}min=1\\ max=9\\ mean=2.1\end{tabular} & \begin{tabular}[c]{@{}c@{}} min=1:5\\ max=5:1 \end{tabular}  \\ \hline
	\end{tabular}
	\vspace{0.2cm}
	\caption{\emph{Visually} dataset statistics. We curated the original 63K infographics available on \emph{Visual.ly} to produce a representative dataset with consistent tags and sufficient instances per tag.} \label{tab:datasetstats}
\end{SCtable}
The infographics in \emph{Visually29K} are very large: up to 5000 pixels per side. Over a third of the infographics are larger than $1000\times1500$ pixels. Aspect ratios  vary between 5:1 and 1:5. 
Visual and textual elements occur at a variety of scales, and resizing the images for visual tasks may not be appropriate, given that smaller design elements may be lost. Recognizing this, we process the infographics by sampling windows from them at a range of scales (Sec.~\ref{sec:icondataset}).

\subsection{Human annotations of icons}\label{ssec:humanannotations}

\begin{SCfigure}
	\includegraphics[width=0.5\linewidth]{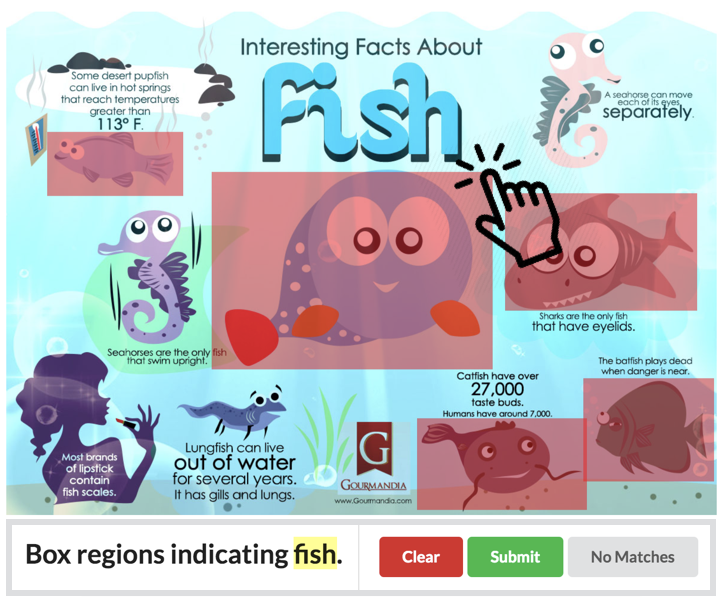}
	\caption{User interface for collecting human ground truth to evaluate icon detection and classification. Participants were either asked to annotate all icons on an infographic, or to only annotate icons corresponding to a particular tag (e.g., \emph{fish}).}
	\label{fig:mturk}
\end{SCfigure}

For a subset of infographics from \emph{Visually29K}, we designed two tasks to collect icon annotations to be used as ground-truth for evaluating computational models (Fig.~\ref{fig:mturk}). In the first task, we asked participants to annotate all the icons on infographics. In the second task, we asked participants to annotate only the icons corresponding to a particular tag. We used the annotations from the first task for evaluating icon proposals (Sec.~\ref{sec:iconnessevaluation}), and the second task for evaluating visual hashtags (Sec.~\ref{sec:userstudies}).

\textbf{Tag-independent icon annotations:} For 1,400 infographics, we asked participants to ``put boxes around any elements that look like icons or pictographs''. No further definitions of ``icon'' were provided. 
This was a time-consuming task, requiring an average of 15 bounding boxes to be annotated per infographic.
A total of 45 participants were recruited, 
producing a total of 21,288 bounding boxes across all 1,400 infographics. 
We split these annotated infographics into 400 for validation (training experiments in Sec.~\ref{ssec:synthetictests}) and 1,000 for testing (reporting final performance in Sec.~\ref{sec:iconnessevaluation}).

\begin{SCfigure}
	\includegraphics[width=0.5\linewidth]{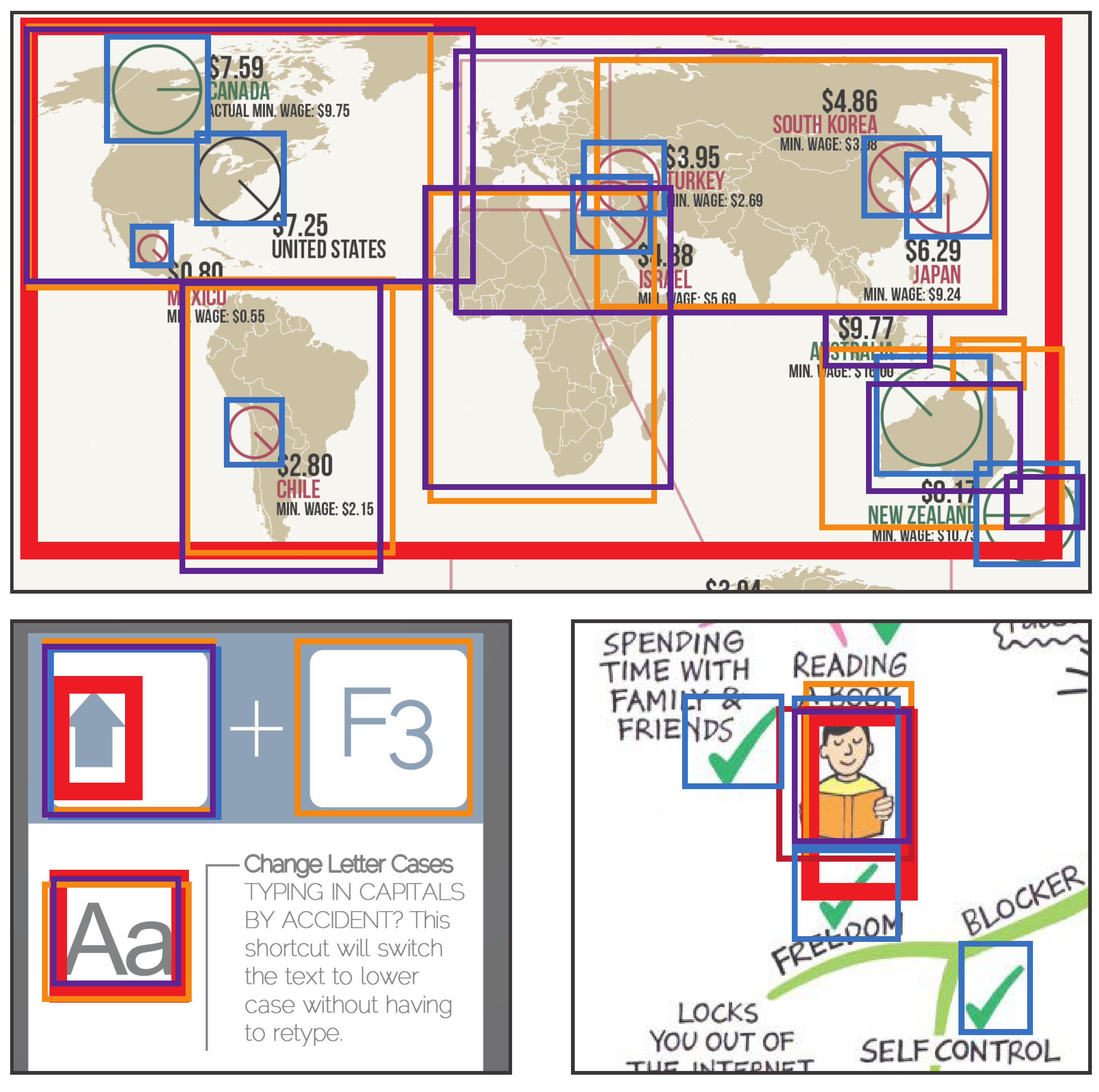}
	\caption{Human agreement in annotating icons is not perfect because different people have different interpretations of ``icon". Here we include 3 crops from annotated infographics. In the world map crop we notice three strategies: (i) labeling the entire map as an icon, (ii) labeling individual continents, (iii) labeling the circle graphics superimposed on the map. The set of participant annotations used for evaluation purposes are indicated in red. The other colored boxes depict annotations from additional participants asked to complete the same task for consistency analyses.} 
	\label{fig:consistencies}
\end{SCfigure}

\textbf{Annotation consistency:} Because the interpretation of ``icon'' may differ across participants, we wanted to measure how consistently humans annotate icons on infographics.
For each of 55 infographics, we recruited an additional 5 annotators. Annotations of these participants were compared to the original collected annotations (above). We use human consistency as an upper bound on computational models. The scores were averaged across participants and images and are reported in Table~\ref{tab:results_iconness}. Human precision and recall are not perfect because different people might disagree about whether a particular visual element (e.g.,~map, embedded graph, photograph) is an icon. They may also disagree when annotating the boundaries of the icon (Fig.~\ref{fig:consistencies}). 

\textbf{Tag-conditional icon annotations:} As ground truth for icon classification, we collected finer-grained annotations by giving participants the same task as before, but asking them to mark bounding boxes around all icons that correspond to a specific text tag (Fig.~\ref{fig:mturk}). We used 544 infographics along with their associated \emph{Visually29K} text tags, 
to produce a total of 1,110 separate annotation tasks (each task corresponding to a single image-tag pair). From all these tasks, participants indicated that for 275 (25\%) there were no icons on the infographic that corresponded to the text tag. 
For the remaining 835 image-tag pairs, we collected a total of 7,761 bounding boxes from 45 undergraduate students, averaging 9 bounding boxes per image-tag pair. To compute human consistency for this task as well, for 55 infographics (a total of 172 image-tag pairs) we got an additional 5 annotators. This human upper bound is reported in Table~\ref{tab:results_hashtags}. 

\vspace{-2em}
\section{A synthetic data approach for training icon proposals}
\label{sec:icondataset}


We use the term \emph{icon} to refer to any visual element that has a well-defined closed boundary in space and different appearance from the background (i.e., can be segmented as a stand-alone element). This is inspired by how an object is defined by Alexe et al.~\cite{alexeobjectness}. Our approach is more related to \emph{objectness} than to object detection in that we are after class-agnostic object proposals: regions in the image containing icons of any class.  
While finding icons can be useful for graphic designs more generally, here we train and test icon proposals on our own dataset of infographics.

Training an object detector often requires a large dataset of annotated instances
, which is a costly manual effort. We took a different approach, leveraging the fact that infographics are digitally-born to generate synthetic training data: 
we augmented existing infographics from the \emph{Visually29K} dataset with Internet-scraped icons. The advantage of this approach is that we can synthesize any amount of training data by repeatedly sampling new windows from infographics and selecting appropriate patches within the windows to paste new icons into.

\vspace{-0.5em}
\subsection{Synthetic dataset creation}

\textbf{Collecting icons:}
Starting with the 391 tags in the \emph{Visually29K} dataset, we queried Google with the search terms `dog icon', `health icon', etc. for each tag. 
The search returned a wide range of stylistically and semantically varied icon images (Fig.~\ref{fig:scraped_icons}).  
We scraped 250K icons with both transparent and non-transparent backgrounds.
Only transparent-background icons were used to augment infographics and train our final icon proposal mechanism, while all 250K icons were used for training an icon classifier (Sec.~\ref{sec:hashtags}). We also present results of training an icon proposal mechanism with icons without transparent backgrounds (Sec.~\ref{sec:iconnessevaluation}). 

\begin{figure}[h]
	\centering
	\includegraphics[width=\linewidth]{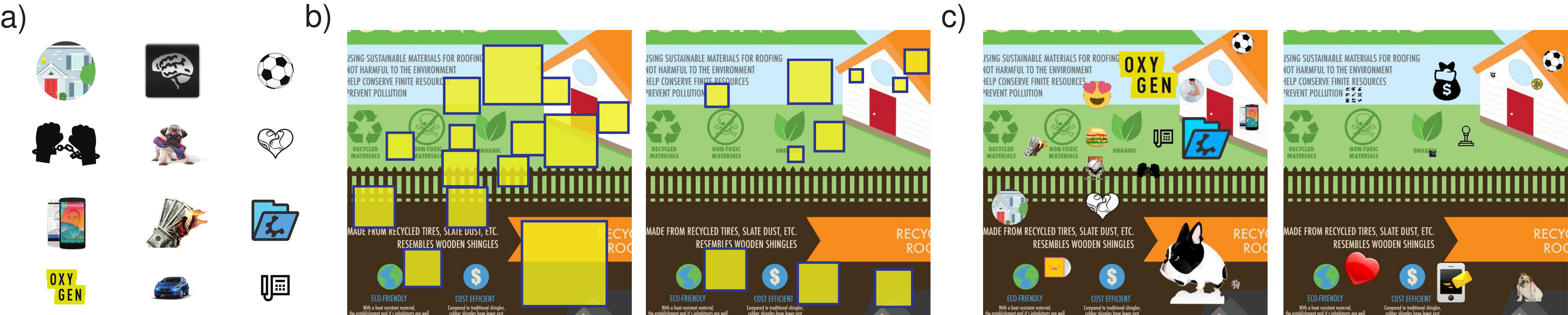}
	\caption{Synthetic data generation pipeline. a) Icons with transparent backgrounds scraped from Google. b) Patches selected for augmenting icons, using different approaches. The approach on the left allows more overlap of icons with background elements. The approach on the right is more conservative, selecting appropriate patches to add icons to. c) Infographic windows augmented with the scraped icons.}
	\label{fig:synthetic_gen}
\end{figure}

\textbf{Augmenting infographics:}
To create our synthetic data, we randomly sampled $600\times600$px windows from the \emph{Visually29K} infographics. Each window was analyzed for patches of low entropy: measuring the amount of texture in a patch to determine if it is sufficiently empty for icon augmentation. Specifically, from a window, a random patch (with varying location and size) was selected, and Canny edge detection was applied to the patch. The resulting edge values were weighted by a Gaussian window centered on the patch, to give more weight to edges in the center of the patch, and summed to quantify the local entropy, with value ranging from 0 to 1. If the entropy value was below a predefined threshold, the patch was kept, otherwise it was discarded and a new patch was sampled from the window (Fig.~\ref{fig:synthetic_gen}b). A randomly selected icon from our scraped icon collection was augmented onto each valid patch in a window, for a fixed number of patches per window (Fig.~\ref{fig:synthetic_gen}c). 
An additional constraint required the icon to meet a set contrast threshold with the patch to ensure it would be visually detectable, or else a new icon would be selected. 

\vspace{-0.5em}
\subsection{Effect of synthetic data parameters on model performance}\label{ssec:synthetictests}
We analyzed how different data augmentation strategies affect the icon proposals on a set of 400 validation infographics containing ground truth annotations of 7,020 bounding boxes. 
We performed a grid search on 4 augmentation parameters, varying them one at a time: (a) number of icons augmented per window, (b) variation in the size of augmented icons, (c) contrast threshold between the icon and the patch: calculated as difference in color variance between the patch and the icon, and (d) entropy threshold for a patch to be considered valid for augmentation. 

We tried 5 settings for the number of icons augmented, from 1 to 16, doubling the number of icons for each experiment. We found no statistically significant differences in the mAP scores of the models trained with these settings. However, increasing the number of icons augmented per patch increases the time required to generate the synthetic data, since we need to find enough valid image patches to paste icons into. We found that higher scale variation during training helps the model detect icons in infographics, which often occur at different scales. By allowing icons to be augmented at sizes ranging from 30 to 480 pixels per side, we achieved the highest mAP scores. Other settings we tried included capping the maximum icon size at 30, 60, 120, and 240 pixels per side. Icons larger than 480 pixels per side were not practical with our $600\times600$px windows.

We found no significant effects of varying the contrast and entropy thresholds independently, while keeping the other augmentation parameters fixed. However, when we disregard both thresholds and place icons entirely at random in the image windows, the performance of the trained model degrades significantly (Sec.~\ref{sec:iconnessevaluation}).  
For generating icon proposals on test images, we chose the model with the highest mAP score on the 400 validation images, with 4 icons per window and icon sizes varying from 30 to 240 pixels per side. 

\section{Learning to propose icons}\label{sec:iconness} 
We can now use our synthetic data to learn to detect icons.
We use the Faster R-CNN network architecture~\cite{FasterRCNN}, although it is worth noting our training procedure with synthetic icon data can be applied with any architecture. Similar to Dwibedi et al.~\cite{dwibedi2017cut}, we are motivated by the fact that Faster R-CNN puts more emphasis on the local visual appearance of an object rather than the global scene layout. As a result, the fact that icons can occur at any location on an infographic is not a problem for the approach.

We adapted Faster R-CNN by making three changes: (i) to handle the large size of infographic images, each image was fed as a cascade of crops, and the detections were aggregated; (ii) we changed the last layer to classify only two categories: any type of icon versus background; (iii) early termination was used during training because the network was found to converge in significantly fewer epochs than the original paper. This could be because detecting generic ``iconic'' regions depends more on low-level information while category-specific details don't need to be learned. 

\textbf{Multi-scale detection at test-time:}
Infographics in the \emph{Visually29K} data-set are large and contain features at different scales. 
At test-time, we sampled windows from infographics at 3 different scales to be fed into the network. The first scale spans the entire image. For the two subsequent scales, we sampled 4 and 9 windows, respectively, such that (i)~windows at each scale jointly cover the entire image, and (ii) neighboring windows overlap by 10\%. Before being fed into the network, every window was rescaled to $600\times600$px. 
The predicted detections from each window were thresholded. Detections across multiple scales were aggregated using non-maximal suppression (NMS) with a value of 0.3. Finally, NMS was applied again to combine smaller detections (often parts of icons) to obtain the final predictions.

\textbf{Training details:} We used a total of 10K training instances (windows), where each window was provided with bounding boxes corresponding to the synthetically augmented icons. Faster R-CNN was trained for 30K iterations with a learning rate of $10^{-3}$. Each iteration used a single augmented window to generate a mini-batch of 300 region proposals. 

\section{Evaluation of icon proposals}\label{sec:iconnessevaluation}

To evaluate our icon proposals, we compare to human annotations of icons on 1,000 test infographics. 
We report performances using standard detection metrics: precision (\emph{Prec}), recall (\emph{Rec}), F-measure, and mAP. To compute precision and recall, we threshold IOU at 0.5 (as in the VOC challenge~\cite{everingham2010pascal}). F-measure is defined as: 
\begin{equation*}
\small
F_\beta = \frac{(1+\beta^2)Prec\times Rec}{\beta^2Prec + Rec}
\end{equation*}
We set $\beta = 0.3$ to weight precision more than recall (a common setting~\cite{borji2015salient}).

Our icon proposal task is related to objectness, general object detection, and object segmentation. We evaluated 5 methods spanning these different tasks, originally trained on natural images, to evaluate the representation gap when applied to infographics. We used objectness~\cite{alexeobjectness}, state-of-the-art object detectors YOLO9000~\cite{YOLO9000}, SSD~\cite{SSD}, and Faster R-CNN~\cite{FasterRCNN}, and class-agnostic object masks~\cite{Sharpmask}. To treat the outputs of the object detectors as object proposals (class-agnostic detections), we report any detection above threshold for any object class that these detectors predict. Default parameters were used for Faster R-CNN and YOLO9000. We report SSD detections with the best setting (thresholded at 0.01). 
Re-training Faster R-CNN with our synthetic data (our full model)
significantly outperformed networks trained on natural images (Table~\ref{tab:results_iconness}). 

\textbf{Evaluation of synthetic data design choices:} To evaluate the contributions of the main design choices in generating our synthetic data, we ran three additional baselines: (a) augmenting icons in random locations on infographics (instead of finding background patches with low entropy and high contrast with icons), (b) augmenting icons without transparent backgrounds, so that when pasted on an infographic, the augmented icons have clearly-visible boundaries, (c) augmenting icons onto white backgrounds, rather than infographic backgrounds. The last baseline is most similar to the approach in Tsutsui and Crandall~\cite{tsutsui2017data}. From Table~\ref{tab:results_iconness} we see that all three baselines perform significantly worse than our full model, demonstrating the importance of all three of our design choices: pasting icons with (i) transparent backgrounds onto (ii) appropriate background patches of (iii)~in-the-wild infographics. We note that the worst performance among the baselines was when icon proposals were not trained with appropriate backgrounds.

\begin{table}[]
\small
\centering
\begin{tabu} to \linewidth {c c c c c c}
\toprule
\textbf{Training data} & \textbf{Model} 				& \textbf{Prec.} 	& \textbf{Rec.}	& \textbf{F$_{0.3}$}      	& \textbf{mAP} \\ \midrule
\multirow{3}{*}{Synthetic with icons} & Full model (ours)						& 38.8	& 34.3	& 43.2	& 44.2 \\ 
& Random locations	& 	29.1 & 15.1	& 29.6	& 32.5  \\ 
& Non transparent icons	& 	24.6 & 17.1	& 25	& 26.1  \\ 
& Blank background	& 7.9	& 24.3	& 10.1	& 10.3  \\ \midrule 
\multirow{5}{*}{Natural images} & YOLO9000~\cite{YOLO9000}	        & 13.6  &  7.1  & 12.6	& 13.7 \\ 
& Faster R-CNN~\cite{FasterRCNN}		& 11.0 	&  6.0	& 10.2	& 11.4 \\ 
& SSD~\cite{SSD}	            		&  9.3  &  34.2 &  10.0	& 11.4 \\
& Objectness~\cite{alexeobjectness}	&  2.9	&  5.6	&  3.1	&  3.0 \\ 
& Sharpmask~\cite{Sharpmask}			&  1.1	&  1.4	&  1.2	&  1.1 \\ \midrule
& Human upper bound								&  63.1	&  64.7	& 61.8	& 66.3 \\ 
\bottomrule
\end{tabu}
\small
\caption{Model performance at localizing icons in infographics. The first 4 models were trained with synthetic data containing icons. The next 5 models were trained to detect objects in natural images. The human upper bound is a measure of human consistency on this task. All values are listed as percentages.}\label{tab:results_iconness}
\end{table}

\vspace{-3em}
\section{Application: multi-modal summarization}\label{sec:hashtags}

\begin{figure}
	\centering
	\includegraphics[width=1\linewidth]{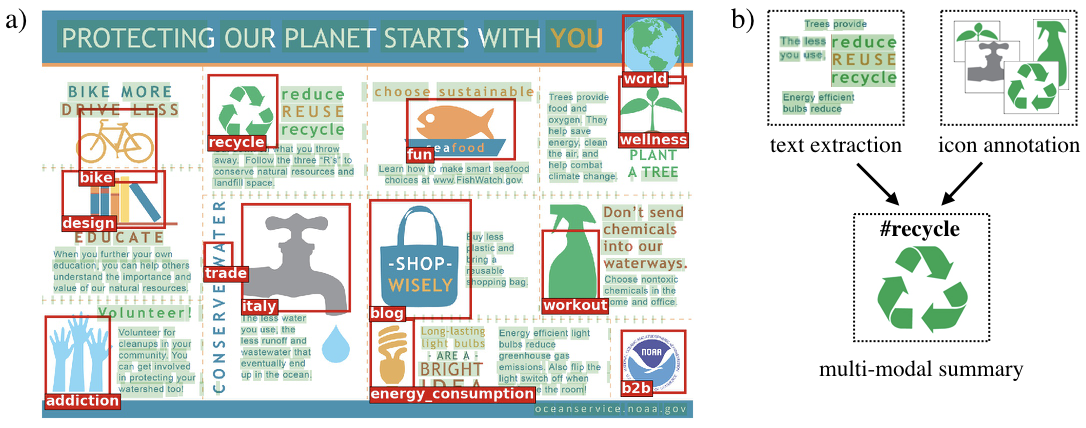}
	\caption{Our computational pipeline for parsing an infographic and computing a multi-modal summary. a) The output of our fully-automatic annotation system, running text detection and OCR using Google's Cloud Vision API \cite{googleText} (semi-transparent green boxes), and our icon detection and classification (red outlines). We trained an icon proposal mechanism with synthetic data to make this system possible. The underlying infographic has been faded to facilitate visualization. b) Our multi-modal summarization application uses the detected text and icons on an infographic to produce the text tags and visual hashtags most representative of the infographic's topics. }
	\label{fig:dense}
\end{figure}

Detecting the textual and visual elements in graphic designs like infographics can facilitate knowledge retrieval, captioning, and summarization applications. As a first step towards infographic understanding, we propose a multi-modal summarization application built upon automatically-detected visual and textual elements (Fig.~\ref{fig:dense}). Just as video thumbnails facilitate the sharing, retrieval, and organization of complex media files, our multi-modal summaries can be used for effectively capturing a visual digest of complex infographics.
Given an infographic as input, our multi-modal summary consists of textual and visual hashtags representative of an infographic's topics. We define \emph{visual hashtags} as icons that are most representative of a particular text tag. 
We evaluate the quality of our multi-modal summary by separately testing each component of the pipeline against a set of human annotations.

\subsection{Approach}

\begin{figure}
	\centering
	\includegraphics[width=1\linewidth]{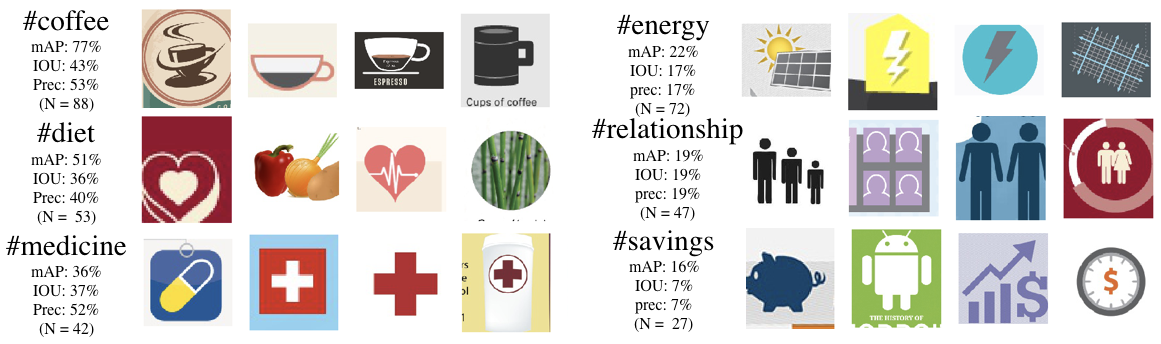}
	\caption{Visual hashtags for different concepts. We include 6 different tag classes, sorted by mAP. For each tag class, depicted are the top 4 instances with highest classifier confidence for each tag, constrained to come from different images. Also indicated is the total number (N) of icon proposals per tag class.}\label{fig:vishashtags}
\end{figure}

\begin{figure*}
	\centering
	\includegraphics[width=1\linewidth]{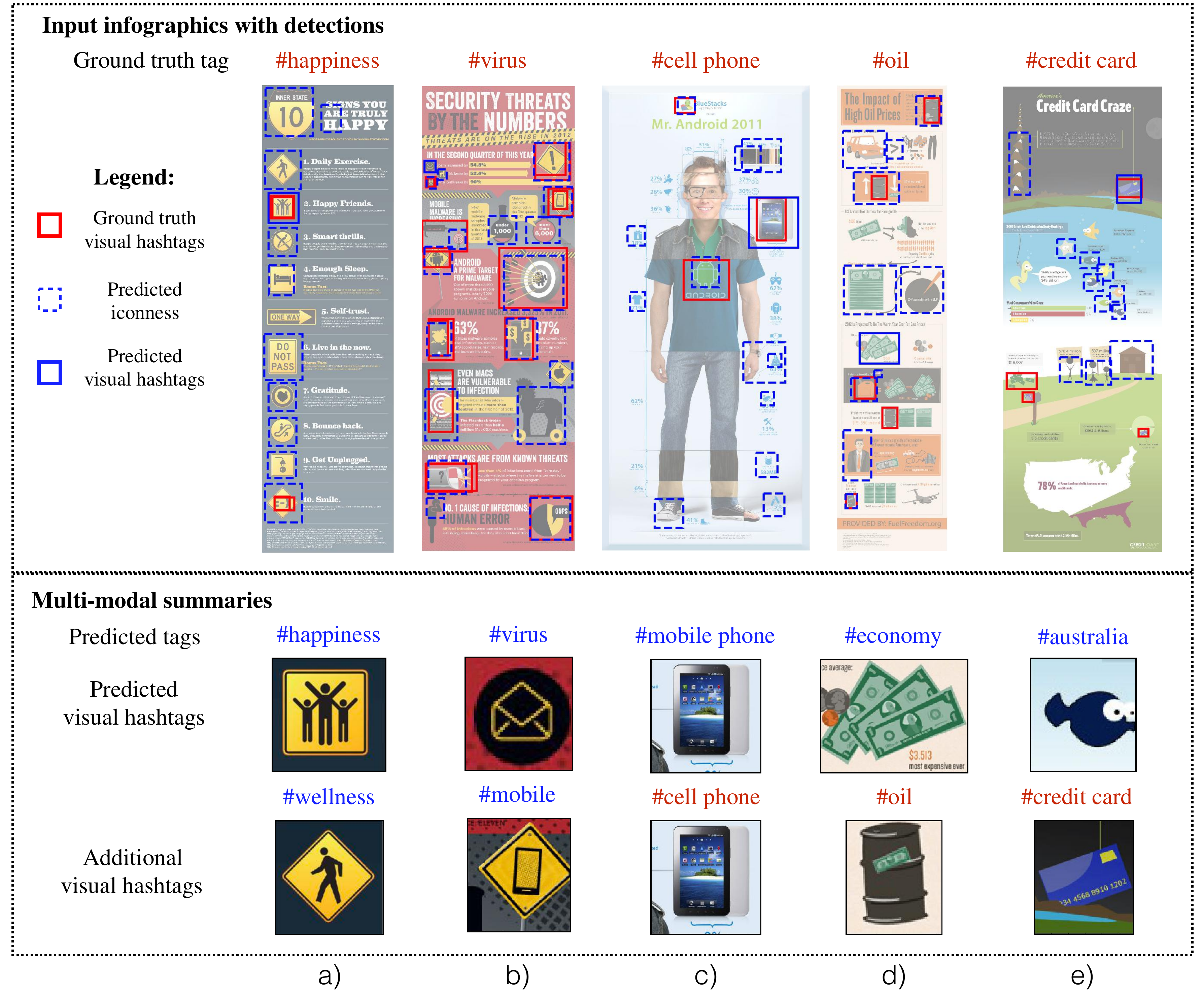}
	\vspace{-1em}
	\caption{Examples of our automated multi-modal summarization pipeline, which given an infographic as input, predicts text tags and corresponding visual hashtags. In both (a) and (b), the predicted text tags for the infographics are correct, and the predicted visual hashtags (solid blue boxes) overlap with human annotations (red boxes). Because a single tag might not be sufficient to summarize an infographic, we also provide an additional predicted text tag (second most likely) and corresponding visual hashtag for (a) and (b). In (c)-(e) the text model predicts the wrong tag. In (c), the semantic meaning of the predicted tag is preserved, so the visual hashtag is still correct. In (d) and (e), the wrong visual hashtags are returned as a result of the text predictions. However, we show that if the correct text tag would have been used (bottom, red), correct visual hashtags would have been returned. In dashed blue are all our icon proposals for each infographic.
		The underlying infographics have been faded to facilitate visualization.}\label{fig:fullpipeline}
\end{figure*}

\textbf{Predicting text tags:} We used Google's Cloud Vision optical character recognition~\cite{googleText} to detect and parse the text from infographics. 
On average, we extracted 236 words per infographic, of which 170 had \emph{word2vec} representations ~\cite{googleWord2VecModel,DBLP:journals/corr/abs-1301-3781}.
The 300-dimensional mean \emph{word2vec} of the bag of extracted words was used as the global feature vector of the text for the infographic.
This feature vector was fed into a single-hidden-layer neural network for tag prediction. Since each infographic could have multiple tags, we set this up as a multi-label problem with 391 tags.\\
\indent \textbf{Classifying icon proposals:} We used the ResNet18 architecture~\cite{Resnet} pre-trained on ImageNet, and fine-tuned on icons scraped from Google along with their associated tags (Sec.~\ref{sec:icondataset}). Training was set up as a multi-class problem with 391 tag classes.
In addition to the icons with transparent backgrounds, icons with non-transparent backgrounds facilitated the generalization of icon classification to automatically-detected icons.\\
\indent \textbf{Predicting visual hashtags:} For an input infographic, we predict text tags and generate icon proposals. All the proposals are then fed to the icon classifier to produce a 391-dimensional feature vector of tag probabilities. 
Then, for each predicted text tag, we return the icon with the highest probability of belonging to that tag class. Fig.~\ref{fig:vishashtags} contains examples of some visual hashtags: the most confident detections for different tag classes. We demonstrate the automatic output of our system in Fig.~\ref{fig:fullpipeline}: given an infographic, we predict the text tag and corresponding visual hashtag.

\subsection{Evaluation}\label{sec:userstudies}

Given an infographic, to evaluate the quality of our predicted text tags, we compared them to the ground truth tags in the \emph{Visually29K} dataset.
To evaluate the ability of our computational system to output a relevant visual hashtag for a given infographic and tag, we compare against the human annotations for 544 \emph{Visually29K} infographics (Sec.~\ref{ssec:humanannotations}). Similar to the task that our computational system receives, participants were asked to annotate all icons corresponding to a particular text tag on an infographic.




\textbf{Evaluation of text tag prediction:} Each infographic in \emph{Visually29K} comes with 1-9 tags (2 on average). 
We achieved 42.6\% top-1 average precision and 24.6\% top-1 average recall at predicting at least one of an infographic's tags. 

\textbf{Evaluation of icon classification:} Before evaluating visual hashtags on a per-infographic basis, we evaluate the ability of the icon classifier to retrieve relevant icons \emph{across infographics}. For each of 391 tags, we used the icon classifier's confidence to re-rank all the icon proposals extracted from 544 infographics. Fig.~\ref{fig:vishashtags} contains the highest confidence icon proposals for a few different tags. For each icon proposal, we measure overlap with human annotations: if an icon proposal sufficiently overlaps with a ground truth bounding box (IOU$>0.5$), that proposal is considered successful. 
We obtained a mAP of 25.1\% by averaging the precision of all the retrieved icon proposals (across all tags).

\textbf{Evaluation of visual hashtags:} Next we evaluate the ability of the classifier to retrieve visual hashtags: icons representative of a particular tag, on a \emph{per-infographic} basis.  
For the following evaluation, we assume that the input is an infographic and a text tag (or multiple tags, if they exist). In a fully automatic setting, the text tags would be predicted by the text model. Fig.~\ref{fig:fullpipeline} contains sample results from the fully automatic pipeline.
Here we evaluate the quality of our proposed visual hashtags independently of the text model's performance. 

For the 835 image-tag pairs with human annotations, we computed the IOU of each of our predicted hashtags with ground truth. 
We evaluated precision as the percent of predicted visual hashtags that have an IOU $> 0.5$ with at least one of the ground truth annotations. 
Human participants may annotate multiple icons for an image-tag pair (Fig.~\ref{fig:fullpipeline}, red boxes). Our application is intended to return a single visual hashtag for a given image-tag pair (Fig.~\ref{fig:fullpipeline}, solid blue boxes), so we report top-1 precision (Table~\ref{tab:results_hashtags}). 
However, we also include the mAP score by considering all our proposals per image-tag pair 
(Fig.~\ref{fig:fullpipeline}, dashed blue boxes). 
From Table~\ref{tab:results_hashtags} we see that sorting the icon proposals using our icon classifier produces more relevant results (mAP = 18.0\%) for a given tag than just returning the most confident (class-agnostic) icon proposals (mAP = 14.5\%). We also verify again that the icon proposals generated by training with icons with \emph{transparent} backgrounds augmented onto \emph{appropriate background regions} of \emph{in-the-wild infographics} outperform baselines that were also trained with synthetic data but with one of these aspects missing.




\begin{SCtable}[]
	\small
	\centering
	\begin{tabu} to \linewidth {c c c c}
		\toprule
		\textbf{Model} 								& \textbf{Top-1 Prec.}  	& \textbf{mAP}		\\ \midrule
		Icon proposals + classification			& 27.2  	&	18.0	\\ 
		Random locations + class.	& 16.7 & 14.2	\\ 
		Non transparent icons + class.	& 15.9 & 14.5	\\ 
		Blank background + class.	& 16.2 & 14.5 \\ 
		Icon proposals					& 16.2	 	& 14.5 \\ \midrule
		Human upper bound								& 55.4		& 57.2	\\
		\bottomrule
	\end{tabu}
	\caption{Given an infographic and text tag as input, we evaluate the visual hashtags returned. For each image-tag pair, we compute IOU with the ground truth bounding box annotations. A successful visual hashtag is one that has an IOU~$>0.5$ with at least one ground truth bounding box. All values are listed as percentages.}\label{tab:results_hashtags}
\end{SCtable}	

\vspace{-2em}
\section{Conclusion}

%
The space of complex visual information beyond natural images has received limited attention in computer vision
, but this is changing with the increasing popularity of work on graphic designs~\cite{zitnick2013bringing,zitnick2016adopting,comicspaper,hussain2017automatic,wilber2017bam,kembhavi2016eccv}. Within this space, we presented 
a novel dataset of infographics, \emph{Visually29K}, containing a rich mix of textual and visual elements. We developed a synthetic data generation methodology for training an icon proposal mechanism. We showed that key design decisions for our synthetic data included augmenting icons with transparent backgrounds onto appropriate background regions of infographics. Our trained icon proposals generalize to real-world infographics, and together with a text parsing system \cite{googleText} and an icon classifier, can be used to annotate infographics. We presented a multi-modal summarization application, which given an infographic as input, produces text tags and visual hashtags to summarize the infographic's topics.


Infographics are specifically designed with a human viewer in mind, characterized by higher-level semantics, such as a story or a message. Beyond simply detecting and classifying the objects contained within them, an understanding of these infographics involves understanding the included text, the layout and spatial relationships between the elements, and the intent of the designer. Human designers are experts at piecing together elements that are cognitively salient (or memorable) and maximize the utility of information. This new space of multi-modal data can give computer vision researchers the opportunity to model and understand the higher-level properties of textual and visual elements in the story being told.

\section{Acknowledgements}
The authors would like to thank Anelise Newman and Nathan Landman for feedback on earlier versions of this manuscript.
%
%
%
%
\bibliographystyle{splncs04}
\bibliography{egbib}

\begin{thebibliography}{10}
\providecommand{\url}[1]{\texttt{#1}}
\providecommand{\urlprefix}{URL }
\providecommand{\doi}[1]{https://doi.org/#1}

\bibitem{alexeobjectness}
Alexe, B., Deselaers, T., Ferrari, V.: Measuring the objectness of image
  windows. IEEE TPAMI  \textbf{34}(11),  2189--2202 (2012)

\bibitem{borji2015salient}
Borji, A., Cheng, M.M., Jiang, H., Li, J.: Salient object detection: A
  benchmark. IEEE TIP  \textbf{24}(12),  5706--5722 (2015)

\bibitem{Sintel}
Butler, D.J., Wulff, J., Stanley, G.B., Black, M.J.: A naturalistic open source
  movie for optical flow evaluation. In: ECCV (2012)

\bibitem{bylinskii2017learning}
Bylinskii, Z., Kim, N.W., O'Donovan, P., Alsheikh, S., Madan, S., Pfister, H.,
  Durand, F., Russell, B., Hertzmann, A.: Learning visual importance for
  graphic designs and data visualizations. In: UIST (2017)

\bibitem{PHAV}
De~Souza, C., Gaidon, A., Cabon, Y., Lopez~Pena, A.: Procedural generation of
  videos to train deep action recognition networks. In: CVPR (2017)

\bibitem{dwibedi2017cut}
Dwibedi, D., Misra, I., Hebert, M.: Cut, paste and learn: Surprisingly easy
  synthesis for instance detection. ICCV  (2017)

\bibitem{everingham2010pascal}
Everingham, M., Van~Gool, L., Williams, C.K., Winn, J., Zisserman, A.: {The
  Pascal Visual Object Classes (VOC) Challenge}. IJCV  \textbf{88}(2),
  303--338 (2010)

\bibitem{VirtualKITTI}
Gaidon, A., Wang, Q., Cabon, Y., Vig, E.: Virtual worlds as proxy for
  multi-object tracking analysis. In: CVPR (2016)

\bibitem{googleText}
Google: {Cloud Vision API}: {Optical} {Character} {Recogition}.
  \url{https://cloud.google.com/vision/} (accessed in October 2017)

\bibitem{googleWord2VecModel}
Google: Word2vec model. \url{https://code.google.com/archive/p/word2vec/}
  (accessed in October 2017)

\bibitem{oxford_textspotter}
Gupta, A., Vedaldi, A., Zisserman, A.: Synthetic data for text localisation in
  natural images. In: CVPR (2016)

\bibitem{Resnet}
{He}, K., {Zhang}, X., {Ren}, S., {Sun}, J.: {Deep Residual Learning for Image
  Recognition}. In: CVPR (2016)

\bibitem{hussain2017automatic}
Hussain, Z., Zhang, M., Zhang, X., Ye, K., Thomas, C., Agha, Z., Ong, N.,
  Kovashka, A.: Automatic understanding of image and video advertisements. In:
  CVPR (2017)

\bibitem{comicspaper}
{Iyyer}, M., {Manjunatha}, V., {Guha}, A., {Vyas}, Y., {Boyd-Graber}, J.,
  {Daum{\'e}}, III, H., {Davis}, L.: {The Amazing Mysteries of the Gutter:
  Drawing Inferences Between Panels in Comic Book Narratives}. In: CVPR (2017)

\bibitem{kembhavi2016eccv}
Kembhavi, A., Salvato, M., Kolve, E., Seo, M., Hajishirzi, H., Farhadi, A.: A
  digram is worth a dozen images. In: ECCV (2016)

\bibitem{SSD}
{Liu}, W., {Anguelov}, D., {Erhan}, D., {Szegedy}, C., {Reed}, S., {Fu}, C.Y.,
  {Berg}, A.C.: {SSD: Single Shot MultiBox Detector}. In: ECCV (2016)

\bibitem{DBLP:journals/corr/abs-1301-3781}
Mikolov, T., Chen, K., Corrado, G., Dean, J.: Efficient estimation of word
  representations in vector space. CoRR  \textbf{abs/1301.3781} (2013),
  \url{http://arxiv.org/abs/1301.3781}

\bibitem{miller1995wordnet}
Miller, G.A.: {WordNet: a lexical database for English}. Communications of the
  ACM  \textbf{38}(11),  39--41 (1995)

\bibitem{Sharpmask}
{Pinheiro}, P.O., {Lin}, T.Y., {Collobert}, R., {Doll{\`a}r}, P.: {Learning to
  Refine Object Segments}. In: ECCV (2016)

\bibitem{2017-reverse-engineering-vis}
Poco, J., Heer, J.: Reverse-engineering visualizations: Recovering visual
  encodings from chart images. Computer Graphics Forum (Proc. EuroVis)  (2017),
  \url{http://idl.cs.washington.edu/papers/reverse-engineering-vis}

\bibitem{YOLO9000}
{Redmon}, J., {Farhadi}, A.: {YOLO9000: Better, Faster, Stronger}. In: CVPR
  (2017)

\bibitem{FasterRCNN}
{Ren}, S., {He}, K., {Girshick}, R., {Sun}, J.: {Faster R-CNN: Towards
  Real-Time Object Detection with Region Proposal Networks}. In: ICCV (2015)

\bibitem{VideoGame}
Richter, S.R., Vineet, V., Roth, S., Koltun, V.: Playing for data: {G}round
  truth from computer games. In: ECCV (2016)

\bibitem{Imagenet}
Russakovsky, O., Deng, J., Su, H., Krause, J., Satheesh, S., Ma, S., Huang, Z.,
  Karpathy, A., Khosla, A., Bernstein, M., Berg, A.C., Fei-Fei, L.: {ImageNet
  Large Scale Visual Recognition Challenge}. International Journal of Computer
  Vision (IJCV)  \textbf{115}(3),  211--252 (2015).
  \doi{10.1007/s11263-015-0816-y}

\bibitem{satat2017object}
Satat, G., Tancik, M., Gupta, O., Heshmat, B., Raskar, R.: Object
  classification through scattering media with deep learning on time resolved
  measurement. Optics Express  \textbf{25}(15),  17466--17479 (2017)

\bibitem{2011-revision}
Savva, M., Kong, N., Chhajta, A., Fei-Fei, L., Agrawala, M., Heer, J.:
  Revision: Automated classification, analysis and redesign of chart images.
  In: UIST (2011), \url{http://idl.cs.washington.edu/papers/revision}

\bibitem{seo2014diagram}
Seo, M.J., Hajishirzi, H., Farhadi, A., Etzioni, O.: Diagram understanding in
  geometry questions. In: AAAI. pp. 2831--2838 (2014)

\bibitem{tsutsui2017data}
Tsutsui, S., Crandall, D.: A data driven approach for compound figure
  separation using convolutional neural networks. ICDAR  (2017)

\bibitem{wilber2017bam}
Wilber, M.J., Fang, C., Jin, H., Hertzmann, A., Collomosse, J., Belongie, S.:
  {BAM! The Behance Artistic Media Dataset for Recognition Beyond Photography}.
  ICCV  (2017)

\bibitem{zitnick2013bringing}
Zitnick, C.L., Parikh, D.: Bringing semantics into focus using visual
  abstraction. In: CVPR (2013)

\bibitem{zitnick2016adopting}
Zitnick, C.L., Vedantam, R., Parikh, D.: Adopting abstract images for semantic
  scene understanding. IEEE TPAMI  \textbf{38}(4),  627--638 (2016)

\end{thebibliography}

\end{document}